# Design and Development of an Automated Contact Angle Tester (ACAT) for Surface Wettability Measurement

[1]Connor Burgess, [1]Kyle Douin, [*1]Amir Kordijazi

[*]amir.kordijazi@maine.edu

[1]Department of Engineering, University of Southern Maine, P.O. Box 9300, Gorham, ME 04038, USA

**Abstract:**

The Automated Contact Angle Tester (ACAT) is a fully integrated robotic work cell developed to automate the measurement of surface wettability on 3D-printed materials. Designed for precision, repeatability, and safety, ACAT addresses the limitations of manual contact angle testing by combining programmable robotics, precise liquid dispensing, and a modular software-hardware architecture. The system is composed of three core subsystems: (1) an electrical system including power, control, and safety circuits compliant with industrial standards such as NEC 70, NFPA 79, and UL 508A; (2) a software control system based on a Raspberry Pi and Python, featuring fault detection, GPIO logic, and operator interfaces; and (3) a mechanical system that includes a 3-axis Cartesian robot, pneumatic actuation, and a precision liquid dispenser enclosed within a safety-certified frame. The ACAT enables high-throughput, automated surface characterization and provides a robust platform for future integration into smart manufacturing and materials discovery workflows. This paper details the design methodology, implementation strategies, and system integration required to develop the ACAT platform.

## 1. Introduction

Laboratory automation continues to transform scientific workflows by enhancing precision, throughput, and reproducibility while minimizing human error [1,2]. In materials science, high-throughput automated experimentation—including synthesis, characterization, and analysis—enables rapid discovery cycles and accelerates materials design by orders of magnitude [3–6]. For instance, autonomous laboratories have demonstrated the ability to synthesize dozens of novel materials within days—a process that would otherwise span months under manual labor [7].

Contact angle measurement, a widely used method for evaluating surface wettability, has traditionally been labor-intensive and susceptible to operator bias [8–10]. Although recent efforts have introduced semi-automated tools—such as image-analysis plug-ins and smartphone-based goniometers—they remain limited in throughput and integration with upstream processes [11–18].

To our knowledge, no complete system currently exists that fully integrates automated sample handling, precise droplet dispensing, real-time contact angle measurement, and safety-compliant enclosure design. The Automated Contact Angle Tester (ACAT) closes this gap by delivering a fully autonomous work cell capable of executing contact angle testing with minimal human intervention. By combining robotic Cartesian motion, pneumatic sample handling, automated dispensing, and image processing under a unified control architecture, ACAT aligns with the goals of high-throughput experimental platforms—supporting rapid, repeatable surface wettability studies vital for materials development.

This paper presents the design and development of the Automated Contact Angle Tester (ACAT) as a fully integrated system for high-throughput and precise surface wettability testing. Section 2 introduces the overall system architecture and design rationale. Section 3 provides a detailed account of the electrical system, including power distribution, safety circuits, and compliance with industrial standards. Section 4 describes the control software implemented on a Raspberry Pi, emphasizing modularity, safety monitoring, and user interface functionality. Section 5 outlines the mechanical and pneumatic subsystems, detailing the Cartesian motion system, automated liquid dispenser, and structural enclosure. Section 6 discusses the full system integration, validation, and future upgrade potential. Finally, Section 7 concludes with reflections on the impact of ACAT on automated experimental methods and potential directions for further development.

## 2. System Architecture

The Automated Contact Angle Tester (ACAT) is a fully integrated system engineered to automate the measurement of contact angles on 3D-printed surfaces with high precision, repeatability, and safety. Its architecture is composed of three interdependent subsystems: the electrical system, the software control system, and the mechanical and pneumatic system. These systems are organized into modular enclosures and components that work in concert to execute an end-to-end automated testing workflow. At a high level, the system includes an AC power enclosure, a main control enclosure, an operator interface station, and the robotic testing cell. These units collectively support the control and operation of a Cartesian robotic arm, a vacuum-based part-handling system, an automated liquid dispenser, and a contact angle tester.

- The electrical system governs power distribution, circuit protection, actuator control, and safety monitoring. It consists of a 120V AC input converted to 24V DC, distributed through fused circuits to motors, sensors, and the main processor. Industrial safety and wiring standards such as NEC 70 [19], NFPA 79 [20], and UL 508A [21] are followed throughout to ensure safe operation and code compliance.
- The software system, implemented on a Raspberry Pi 4 Model B, acts as the central controller. Written in Python, the software coordinates all automation tasks—motion control, liquid dispensing, fault detection, and operator input—via GPIO logic. The system supports both local and remote control interfaces, allowing flexible deployment in research and production environments.
- The mechanical and pneumatic system provides the physical structure and motion platform necessary for testing. It includes a rigid aluminum-frame enclosure with safety shielding, a 3-axis Cartesian robot, an automated liquid dispensing assembly, and a vacuum-powered end effector. Pneumatic components manage the Z-axis motion, end effector actuation, and dispensing valve operation.

This tripartite design structure ensures a modular, maintainable, and scalable system architecture. The rationale for this design is rooted in several key principles:

1. Functional Modularity: Dividing the system into discrete electrical, software, and mechanical domains improves troubleshooting, simplifies maintenance, and enables independent upgrades to each subsystem.
2. Safety and Compliance: Adherence to international electrical and mechanical safety standards ensures that the system can be safely operated in academic and industrial settings.
3. Repeatability and Throughput: The integration of robotics and automation significantly reduces human variability and enables high-throughput surface wettability testing—critical for modern materials research and development.

The subsequent sections of this paper provide a detailed description of each subsystem. Section 3 discusses the electrical design and compliance features; Section 4 presents the software architecture, control logic, and user interface; and Section 5 describes the mechanical and pneumatic components, including actuation, frame construction, and liquid dispensing.

## 3. Electrical Design

The electrical design of the ACAT was developed to deliver reliable power distribution, control, and safety functionality in compliance with recognized industry standards. The system architecture comprises three main subsystems: the power circuit, the safety circuit, and the control circuit. All designs conform to applicable codes and standards, including NFPA 70 (National Electrical Code)[19], NFPA 79 (Electrical Standard for Industrial Machinery)[20], UL 508A (Standard for Industrial Control Panels)[21], NEMA (National Electrical Manufacturers Association) enclosure ratings[22], IEC 60204-1 (Safety of Machinery – Electrical Equipment)[23], and ISO 13485 (Quality Management Systems for Medical Devices)[24].

### 3.1 Electrical Documentation and Schematics

The complete electrical design is represented through a detailed set of electrical schematics (Appendix I), which include sections such as the title page, index, bill of materials, AC enclosure design, main enclosure distribution, safety circuit, control circuit, operator station wiring, enclosure layouts, and panel cutouts. Spare pages were incorporated between schematic sections to allow for future modifications without requiring restructuring of existing diagrams—a common practice in industrial drafting.

### 3.2 Power Circuit

The power circuit is based on a 120V AC single-phase input at 60 Hz, which matches the standard supply available at the USM John Mitchell Center, where the system is installed. In accordance with NFPA 79 guidelines, live components operating above 50V AC are housed in a lockable enclosure. A fused disconnect switch is installed in the AC enclosure (**Figure 1 (a)** and **(b)**) to prevent accidental energization and to comply with Article 110.25 of the NEC.

The AC enclosure converts 120V AC to 24V DC, which is the operating voltage for most of the system's components. To maintain electrical segregation and ensure safety, the design uses two separate enclosures—one for high-voltage AC and one for low-voltage DC. Each enclosure includes a nameplate detailing the voltage rating, current rating, frequency, phase information, power rating, manufacturer information, serial number, short circuit current rating, enclosure type, and operating temperature.

A Class J slow-blow fuse, rated at 4 amps, was selected to accommodate a 3-amp load with consideration for inrush current, adhering to the NFPA 79 requirement of 125% fuse rating relative to the load. The AC enclosure feeds the main control enclosure with the necessary 24V DC power.

### 3.3 Main Enclosure Distribution

The main enclosure (**Figure 1 (c)**) receives 24V DC from the AC enclosure and distributes power to individual subsystems through branch circuits protected by appropriately rated fuses. These subsystems include the stepper motor drives, the main processor (Raspberry Pi 4B), panel ventilation fans, and the contact angle tester. A 24V common bus is used across schematics (Appendix I) for consistency and simplification.

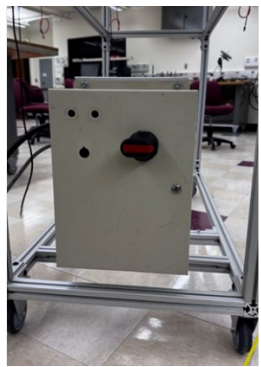
(a)

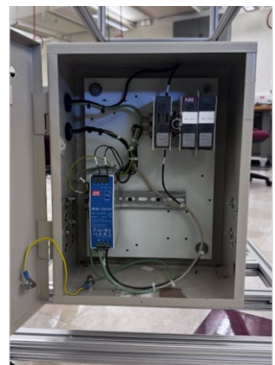
(b)

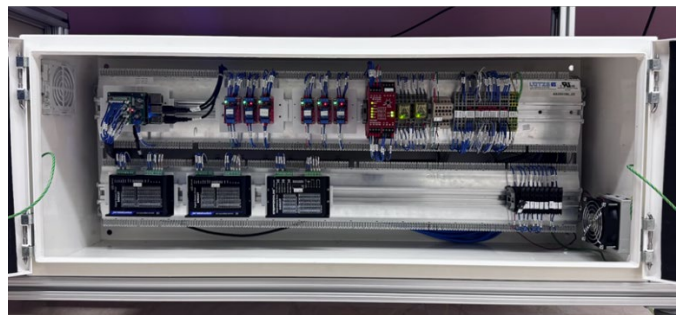
(c)

**Figure 1**. (a) AC enclosure, (b) AC enclosure wiring, (c) Main enclosure wiring

### 3.4 Safety Circuit

The safety system incorporates dual-channel monitoring via a safety relay, which governs the status of emergency stop pushbuttons and a door interlock switch. These elements are configured to detect signal edges; when a fault is detected (e.g., emergency stop activation or door opening), the safety relay disables power to critical components, including the stepper motor drives and pneumatic solenoids. The system achieves a Performance Level (PL) C safety rating through dual redundancy, which exceeds the required PL B rating for this application. A visual indicator (red light) is activated in fault conditions to alert the operator.

### 3.5 Control Circuit

#### 3.5.1 Stepper Motor Drives

Stepper motors used for controlling the Cartesian robot's X, Y, and Z axes are driven via stepper drivers. These are supplied with 24V DC through fused branches monitored by the safety relay. The Raspberry Pi sends PWM (Pulse Width Modulation)-based pulse and direction signals to each drive via twisted-pair shielded cables, which help reduce electrical noise. The system uses NEMA 17 stepper motors with a resolution of 200 steps per revolution. A typical operating frequency of 400 Hz was chosen for drive signaling.

#### 3.5.2 Solenoids and Indicators

The pneumatic actuators and status indicators (light tower) are controlled via 24V relays, triggered by 5V logic-level outputs from the Raspberry Pi. Voltage level isolation is maintained between the control and load circuits. The emergency stop indicator is tied directly to the safety relay, while other indicators are driven by the processor.

#### 3.5.3 Water Dispensing Subsystem

The water dispensing mechanism also uses a stepper motor driven by a stepper driver, with similar configuration to the axis drives. A diaphragm pump, controlled via a relay, supplies water to the dispenser. Both the stepper drive and the pump relay are connected to the processor and monitored by the safety relay to ensure safe operation.

### 3.6 Processor Configuration

The Raspberry Pi 4B serves as the system's main processor and handles all signal processing, I/O control, and automation logic. It is equipped with two HATs (Hardware Attached on Top) to expand GPIO capabilities. The GPIO pins are broken out across four terminal blocks—two for inputs and two for outputs. Each terminal block includes connections to 3.3V or 5V logic power and DC common (ground).

GPIO assignments follow a structured labeling convention. Inputs and outputs are labeled with a prefix (I or O), a block number, and the original GPIO number (e.g., I:1/2). The Raspberry Pi is configured with NPN sinking inputs and PNP sourcing outputs. Input terminals monitor proximity sensors, limit switches, process control buttons (start/stop), and safety faults. Output terminals control stepper drivers and relays for solenoids and indicators.

### 3.7 Operator Control Enclosure

A secondary operator enclosure houses the start/stop buttons and a mushroom-style emergency stop button. The enclosure is connected via shielded cable to reduce electromagnetic interference. The pilot lights for start and stop states are controlled by the Raspberry Pi, while the emergency stop signal is routed directly to the safety circuit.

### 3.8 Panel Layouts and Cutouts

Mechanical drawings for the panel layouts and cutouts were created for the AC enclosure, main control enclosure, and operator interface. These drawings include scaling information for accurate physical construction. As of this writing, some layouts remain in progress, pending final component placement decisions. It is standard drafting practice to complete these drawings once all design constraints and components are finalized to avoid unnecessary redesigns.

## 4. Software Design

The software architecture of the ACAT is implemented on a Raspberry Pi 4 Model B, which serves as the system's main processor. The Raspberry Pi was selected due to its affordability, open-source ecosystem, robust input/output (I/O) capabilities, and extensive community support. Unlike traditional microcontrollers, the Raspberry Pi functions as a full-fledged Linux-based computer, providing advanced connectivity and programming flexibility for automation tasks.

### 4.1 System Interface and Connectivity

The Raspberry Pi is connected to a local operator interface through an HDMI port, allowing the user to interact with the system via a monitor and peripheral devices (e.g., keyboard, mouse). In addition to local access, Secure Shell (SSH) is enabled to allow remote system monitoring, debugging, and software updates. This dual-mode accessibility enhances system flexibility and maintainability in both development and operational phases.

### 4.2 Programming Environment

The control logic is written in Python and developed using the Thonny Integrated Development Environment (IDE), which is optimized for Raspberry Pi environments. Python was chosen due to its ease of use, extensive library support, and ability to interface with GPIO (General Purpose Input/Output) pins for real-time control. The modular and readable structure of Python also facilitates collaboration, documentation, and future expansion of system capabilities.

### 4.3 Control Logic Architecture

The control program is structured into a main process and several modular subprocesses, executed in a sequential and conditional logic flow. Upon system startup, the software continuously monitors the safety status of the hardware. If any fault condition is detected—such as activation of emergency stop buttons or a door interlock breach—the system halts all operations immediately to ensure operator safety and protect hardware components.

Once safety conditions are confirmed and the process is initiated, the software sets all system parameters to their initial conditions and enters the main operational loop. The control system waits for a signal from the operator via the start button. If the stop button is pressed at any time, the loop is interrupted, and the system safely terminates ongoing tasks.

### 4.4 Testing Workflow

Following operator initiation, the ACAT system performs the following automated sequence:

1. Axis Homing: All three Cartesian axes (X, Y, and Z) are homed to establish a reference position.
2. Part Handling: The first test part is picked up using a pneumatic suction system and positioned for testing.
3. Contact Angle Measurement: The water dispenser deposits a droplet on the surface, and the contact angle tester records the measurement.
4. Part Unloading: The tested part is removed from the testing station and placed in an unloading area.

This cycle is governed by a nested loop structure. The outer loop iterates through the columns of test samples, while the inner loop processes each surface within a given column. Once all surfaces in a column are tested, the outer loop advances to the next column. The process continues until all parts (i.e. 25 parts) have been evaluated or until an operator intervention stops the system.

### 4.5 Fault Handling and Termination

Throughout the operation, fault monitoring remains active. If a safety device is triggered at any stage, the software immediately exits the active loop, deactivates motion and dispensing subsystems, and returns the system to a safe state. This integration of software-level safety logic complements the hardware-based safety circuits, providing a robust fault-tolerant design.

**Figure 2**, shows the flowchart of the software control logic for the ACAT.

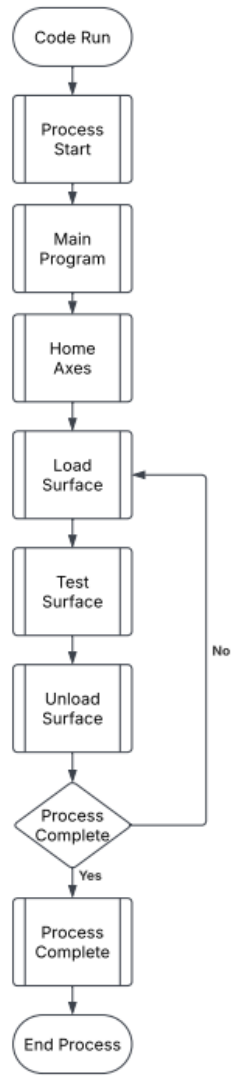

**Figure 2**. ACAT Programming overview flowchart

## 5. Mechanical Design

The mechanical design of the ACAT plays a central role in ensuring system stability, precision, repeatability, and safety during automated contact angle measurement. The mechanical framework supports all critical subsystems—including motion control, sample handling, dispensing, and safety enclosures—while adhering to relevant international engineering standards for industrial automation and laboratory equipment.

### 5.1 Structural Framework and Enclosure

The ACAT system is housed within a custom-built enclosure constructed from 40×40 mm metric aluminum extrusion profiles (**Figure 3 (a)**). This choice of material offers a balance of mechanical rigidity, modularity, and ease of assembly, while maintaining compatibility with metric-standard components. The enclosure dimensions are 147.5 cm in length, 65 cm in width, and 168 cm in height (including wheels), offering a compact yet spacious footprint suitable for benchtop operation.

To enable safe transport and mobility, lockable caster wheels are mounted at the base. Additional crossbeams are strategically positioned beneath the working surface to enhance structural rigidity, reduce vibration during axis movements, and maintain alignment for repeatable measurements. The entire frame is enclosed using 3.5 mm thick transparent acrylic panels on all sides and top, forming a controlled environment that isolates the testing area from external contaminants and airflow disturbances—both of which can significantly affect droplet shape and contact angle measurements. The enclosure also acts as a physical barrier to prevent operator exposure to moving parts.

### 5.2 Compliance with Safety Standards

The mechanical design conforms to a range of international safety and quality standards. Industrial robot safety is addressed through adherence to ISO 10218-1 and ANSI/RIA R15.06, resulting in a fully enclosed workspace with integrated interlocks. Risk assessment and reduction practices are guided by ISO 12100, which informed structural reinforcements and secure component mounting. Additionally, ISO 14644-8 informed decisions regarding enclosure sealing and air quality control to support precision surface testing. Operator safety distances, as defined in ISO 13857, are enforced through panel design and access limitations.

### 5.3 Cartesian Motion System

Automated sample handling is achieved using a three-axis Cartesian robot system, which provides linear motion in the X, Y, and Z directions (**Figure 3 (b)**). The Y-axis spans 136 cm and is mounted to the lateral aluminum profiles. It is belt-driven and powered by a stepper motor, with a custom aluminum and guide rail assembly to enhance stability and minimize deflection. The X-axis is mounted orthogonally to the Y-axis and uses a stepper motor with a ball screw drive for improved accuracy and load-bearing capability.

The Z-axis, which is mounted perpendicularly to the X-axis via a 3D-printed bracket, is pneumatically actuated. The vertical motion enables the placement and retrieval of test samples using an end-effector. This axis is designed to be lightweight, reliable, and responsive, offering high repeatability in vertical positioning.

### 5.4 Loading Station and Electrical Integration

To streamline sample management, a custom loading station was developed using 3D-printed legs and a laser-cut acrylic top plate (**Figure 3 (c)**). The station provides indexed sample slots to ensure consistent placement of test surfaces for robotic pickup. Its modular design allows easy removal and reloading, minimizing operator exposure and enabling batch testing.

Additionally, custom enclosures fabricated from 40×40 aluminum profiles are mounted beneath the work surface to house electrical components. These compartments separate high- and low-voltage systems and provide safe, organized access for maintenance while avoiding interference with moving parts.

### 5.5 Automated Liquid Dispensing System

Precise and consistent droplet deposition is critical for reliable contact angle measurement. To eliminate the inconsistencies associated with manual droplet dispensing, the ACAT incorporates a fully automated liquid dispensing system mounted on a linear guide (**Figure 3 (d)**). The dispenser travels along a dedicated axis to position itself accurately above the test surface. Once in position, it dispenses a controlled 10-microliter droplet before retracting to avoid interfering with subsequent Z-axis movements.

The dispensing valve, a MY2626 pneumatic dispenser, features micrometer-adjustable flow control and an anti-drip mechanism, ensuring clean and consistent droplet formation. The dispenser is mounted to a 3D-printed bracket tailored to the system's geometry. Water is supplied from a vertically mounted reservoir via a Uxcell DC 12V 1300 mL/min diaphragm pump, which offers a compact footprint and reliable performance for low-volume liquid delivery. This closed-loop system eliminates variability in drop volume and placement, improving test repeatability.

### 5.6 Pneumatic Subsystems

Pneumatics are central to several ACAT operations, including Z-axis actuation, water dispensing, and vacuum-based sample handling. A pneumatic control block serves as the centralized hub for managing compressed air distribution to each of these subsystems. The control block integrates solenoid valves triggered by the main processor to route air based on real-time instructions.

The Z-axis uses a pneumatic actuator for vertical motion, selected for its low weight and rapid response compared to motor-driven alternatives. The vacuum-powered end-effector is driven by a venturi-based vacuum generator, which creates negative pressure to securely lift and place test samples without physical clamping. Once the sample is positioned, the vacuum is released to allow clean placement on the contact angle tester.

The dispensing valve is also actuated pneumatically. Controlled bursts of air are applied to the valve to release droplets with high precision. The pressure and timing are regulated to ensure uniform volume output, further enhancing measurement consistency.

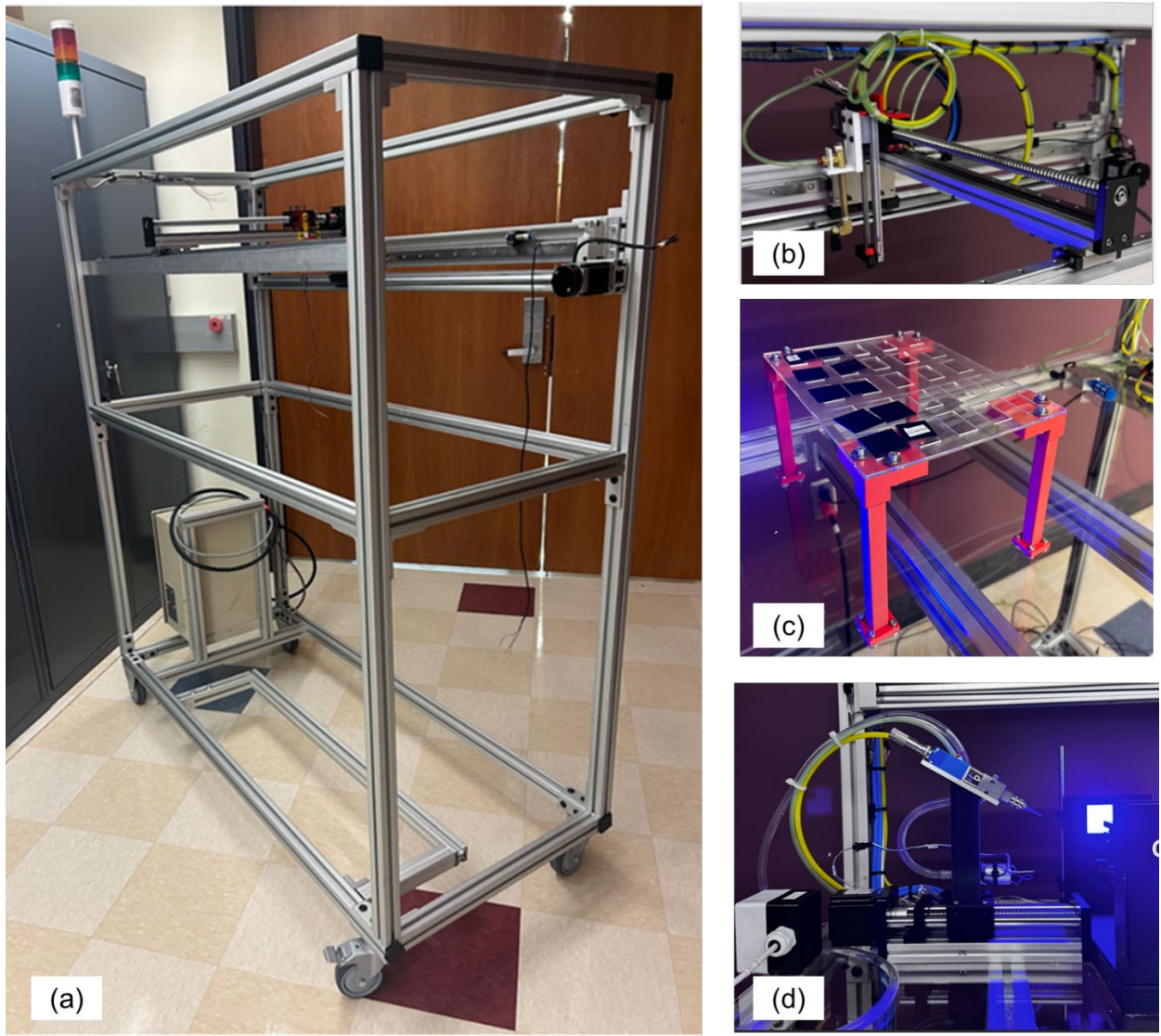

**Figure 3**. (a) Image of the 40/40 aluminum profile, (b) X, Y, and Z cartesian axes, (c) Loading station made of 3D printed parts and acrylic laser cut top, (d) Liquid dispenser, linear axis, and pump holder

## 6. System Integration and Future Directions

The Automated Contact Angle Tester (ACAT) has been successfully developed (**Figure 4 (a)**) as a fully integrated system combining mechanical, electrical, pneumatic, and software components. The complete work cell is designed to perform high-throughput, repeatable, and safe contact angle measurements on a variety of surfaces, with particular emphasis on testing 3D-printed materials. All subsystems have been built, tested, and validated to ensure performance, reliability, and compliance with relevant industrial standards.

The mechanical assembly, based on a 40×40 mm aluminum extrusion frame, provides a rigid and modular structure. The three-axis Cartesian motion platform, including stepper-driven X and Y axes and a pneumatically actuated Z-axis, demonstrated consistent performance in part handling and positioning. The Z-axis was custom-integrated with a vacuum-powered end effector, enabling accurate and damage-free sample transfer. The 3D-printed modular loading station enhances part throughput and supports batch testing, aligning with the system's automation goals.

The electrical system is fully operational and compliant with standards such as NFPA 79, UL 508A, and ISO 13485. It is housed across three enclosures—the AC Enclosure, Main Enclosure, and Operator Control Enclosure—each designed for safety, maintainability, and clarity of function. Comprehensive electrical schematics (Appendix I) document the design and support future troubleshooting and upgrades.

The control software, written in Python and deployed on a Raspberry Pi 4 Model B, was developed with modularity and ease of maintenance in mind. Code organization follows a clear structure, starting with library imports and GPIO pin definitions, followed by functional blocks for specific operations, and culminating in the main control loop that governs the automation sequence. Each GPIO pin is annotated and assigned logically to streamline debugging and future modifications. The software supports both local and remote operation via HDMI and SSH connections, respectively.

Together, these components result in a reliable, safe, and fully autonomous system for contact angle measurement (**Figure 4 (b)**). The ACAT system has been validated to function as intended, demonstrating robust automation of a process that is traditionally manual and error-prone. Its modular design, documented architecture, and standards compliance make it an ideal candidate for further development and deployment in laboratory and industrial environments.

Future enhancements may include integration with programmable logic controllers (PLCs) for industrial-grade control, testing under hazardous or sterile conditions, incorporation into manufacturing lines, and expanded software capabilities such as data logging, machine vision, and remote process monitoring. The ACAT system provides a versatile platform for ongoing innovation in surface characterization and automated materials testing.

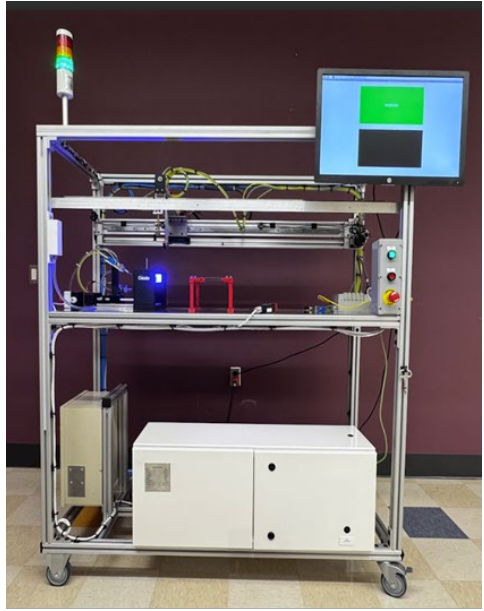
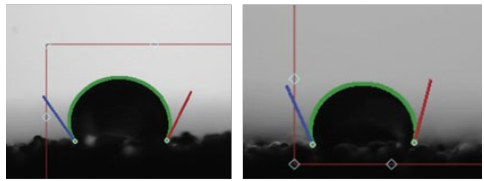

**Figure 4**. (a) Final integrated setup of the Automated Contact Angle Tester (ACAT) system, (b) Representative contact angle measurement obtained using the ACAT platform

## 7. Conclusion

The development of the Automated Contact Angle Tester (ACAT) demonstrates how the integration of automation technologies can significantly enhance the precision, consistency, and efficiency of surface characterization processes. By addressing the limitations of manual contact angle measurement—namely human error, variability in droplet deposition, and positioning inconsistencies—this system offers a robust, repeatable, and scalable solution for material surface analysis.

The ACAT combines a Cartesian robotic system, pneumatic actuation, an automated micro-volume liquid dispenser, and a software interface to execute complex and repetitive tasks with minimal operator intervention. Each subsystem was carefully designed and integrated, drawing upon key principles in mechanical design, control systems, embedded programming, and fluid dynamics. The architecture reflects a modular and maintainable design philosophy, allowing for future enhancements and seamless troubleshooting.

A critical element of the project was strict adherence to relevant industry and safety standards, including NEC 70, NFPA 79, UL 508A, ISO 13485, ISO 12100[25], ISO 10218-1[26], ANSI/RIA R15.06[27], and others governing industrial machinery and laboratory equipment. This compliance ensures that the ACAT system is not only functional but also safe and suitable for deployment in both research and industrial environments.

In its current form, the ACAT provides a fully automated platform for high-throughput contact angle testing. Its design supports adaptability to a range of materials and surface types, and its modular control architecture enables rapid iteration and upgrade. Future work may explore integration with PLCs, in-line manufacturing systems, hazardous material handling protocols, and expanded data analysis capabilities. Additionally, ACAT establishes a foundation for the next generation of laboratory automation systems. Its programmable architecture and autonomous operation make it well-suited for integration into self-driving laboratories, where artificial intelligence algorithms guide experimentation to accelerate surface design and optimization. In such frameworks, ACAT could function as a closed-loop actuator, feeding experimental results directly into AI models to iteratively refine surface treatments and materials for enhanced performance. Ultimately, this project highlights the value of interdisciplinary engineering and automation in advancing experimental methods in materials science and surface engineering.

**Acknowledgment:** This research was supported by the Maine Economic Improvement Fund (award number 6250296) and the Maine Space Grant Consortium (award number 6410247). The authors gratefully acknowledge this support. Special thanks are extended to Corning Life Sciences, Lanco Integrated, and Burgess Controls for their generous donations of electrical components, which were essential to the development of the ACAT system.

**Conflict of Interest:** The authors declare that they have no conflict of interest.

**Data Availability Statement:** The code and integration scripts supporting the ACAT system are available in a public GitHub repository at https://github.com/akordijazi/ACAT.

**References**:

(1) Stricker, M.; Banko, L.; Sarazin, N.; Siemer, N.; Janssen, J.; Zhang, L.; Neugebauer, J.; Ludwig, A. Computationally Accelerated Experimental Materials Characterization -- Drawing Inspiration from High-Throughput Simulation Workflows. arXiv January 27, 2025. https://doi.org/10.48550/arXiv.2212.04804.

(2) Barber, G. Google DeepMind's AI Dreamed Up 380,000 New Materials. The Next Challenge Is Making Them. *Wired*. https://www.wired.com/story/an-ai-dreamed-up-380000-new-materials-the-next-challenge-is-making-them/ (accessed 2025-07-01).

(3) Correa-Baena, J. P.; Hippalgaonkar, K.; Duren, J. van; Jaffer, S.; Chandrasekhar, V. R.; Stevanovic, V.; Wadia, C.; Guha, S.; Buonassisi, T. Accelerating Materials Development via Automation, Machine Learning, and High-Performance Computing. *Joule* **2018**, *2* (8), 1410–1420. https://doi.org/10.1016/j.joule.2018.05.009.

(4) Rosado-Miranda, A. E.; Posligua, V.; Sanz, J. Fdez.; Márquez, A. M.; Nath, P.; Plata, J. J. Design Principles Guided by DFT Calculations and High-Throughput Frameworks for the Discovery of New Diamond-like Chalcogenide Thermoelectric Materials. *ACS Appl. Mater. Interfaces* **2024**, *16* (22), 28590–28598. https://doi.org/10.1021/acsami.4c04120.

(5) Maffettone, P. M.; Banko, L.; Cui, P.; Lysogorskiy, Y.; Little, M. A.; Olds, D.; Ludwig, A.; Cooper, A. I. Crystallography Companion Agent for High-Throughput Materials Discovery. *Nat Comput Sci* **2021**, *1* (4), 290–297. https://doi.org/10.1038/s43588-021-00059-2.

(6) Dippo, O. F.; Kaufmann, K. R.; Vecchio, K. S. High-Throughput Rapid Experimental Alloy Development (HT-READ). arXiv February 11, 2021. https://doi.org/10.48550/arXiv.2102.06180.

(7) Fei, Y.; Rendy, B.; Kumar, R.; Dartsi, O.; P. Sahasrabuddhe, H.; J. McDermott, M.; Wang, Z.; J. Szymanski, N.; N. Walters, L.; Milsted, D.; Zeng, Y.; Jain, A.; Ceder, G. AlabOS: A Python-Based Reconfigurable Workflow Management Framework for Autonomous Laboratories. *Digital Discovery* **2024**, *3* (11), 2275–2288. https://doi.org/10.1039/D4DD00129J.

(8) Kordijazi, A.; Kumar Behera, S.; Suri, S.; Wang, Z.; Povolo, M.; Salowitz, N.; Rohatgi, P. Data-Driven Modeling of Wetting Angle and Corrosion Resistance of Hypereutectic Cast Aluminum-Silicon Alloys Based on Physical and Chemical Properties of Surface. *Surfaces and Interfaces* **2020**, 100549. https://doi.org/10.1016/j.surfin.2020.100549.

(9) Kordijazi, A.; Behera, S.; Patel, D.; Rohatgi, P.; Nosonovsky, M. Predictive Analysis of Wettability of Al–Si Based Multiphase Alloys and Aluminum Matrix Composites by Machine Learning and Physical Modeling. *Langmuir* **2021**. https://doi.org/10.1021/acs.langmuir.1c00358.

(10) Kordijazi, A.; Behera, S. K.; Jamet, A.; Fernández-Calvo, A. I.; Rohatgi, P. Predictive Analysis of Water Wettability and Corrosion Resistance of Secondary AlSi10MnMg(Fe) Alloy Manufactured by Vacuum-Assisted High Pressure Die Casting. *Inter Metalcast* **2025**, *19* (1), 75–85. https://doi.org/10.1007/s40962-024-01327-3.

(11) Chen, H.; Muros-Cobos, J. L.; Amirfazli, A. Contact Angle Measurement with a Smartphone. *Review of Scientific Instruments* **2018**, *89* (3), 035117. https://doi.org/10.1063/1.5022370.

(12) Williams, D. L.; Kuhn, A. T.; Amann, M. A.; Hausinger, M. B.; Konarik, M. M.; Nesselrode, E. I. Computerised Measurement of Contact Angles. *Galvanotechnik* **2010**, *101* (11), 2502.

(13) Akbari, R.; Antonini, C. Contact Angle Measurements: From Existing Methods to an Open-Source Tool. *Advances in Colloid and Interface Science* **2021**, *294*, 102470. https://doi.org/10.1016/j.cis.2021.102470.

(14) Kirk, S.; Strobel, M.; Lyons, C. S.; Janis, S. A Statistical Comparison of Contact Angle Measurement Methods. *Journal of Adhesion Science and Technology* **2019**, *33* (16), 1758–1769. https://doi.org/10.1080/01694243.2019.1611400.

(15) Zou, Y.; Ross, N.; Nawaj, W.; Borguet, E. A Simplified Approach for Dynamic Contact Angle Measurements. *J. Chem. Educ.* **2024**, *101* (9), 3883–3890. https://doi.org/10.1021/acs.jchemed.4c00146.

(16) Saulick, Y.; Lourenço, S. d. n.; Baudet, B. a. A Semi-Automated Technique for Repeatable and Reproducible Contact Angle Measurements in Granular Materials Using the Sessile

- Drop Method. *Soil Science Society of America Journal* **2017**, *81* (2), 241–249. https://doi.org/10.2136/sssaj2016.04.0131.
(17) Nežerka, V.; Somr, M.; Trejbal, J. Contact Angle Measurement Tool Based on Image Analysis. *Exp Tech* **2018**, *42* (3), 271–278. https://doi.org/10.1007/s40799-017-0231-0.
(18) Heiskanen, V.; Marjanen, K.; Kallio, P. Machine Vision Based Measurement of Dynamic Contact Angles in Microchannel Flows. *J Bionic Eng* **2008**, *5* (4), 282–290. https://doi.org/10.1016/S1672-6529(08)60172-9.
(19) *NFPA 70 (NEC) Code Development*. https://www.nfpa.org/codes-and-standards/nfpa-70-standard-development/70 (accessed 2025-07-16).
(20) *NFPA 79 Standard Development*. https://www.nfpa.org/codes-and-standards/nfpa-79-standard-development/79 (accessed 2025-07-16).
(21) *UL 508A Third Edition Summary of Requirements*. UL Solutions. https://www.ul.com/resources/ul-508a-third-edition-summary-requirements (accessed 2025-07-16).
(22) *NEMA Ratings for Enclosures - NEMA Rated Enclosures | Nema Enclosures*. https://www.nemaenclosures.com/enclosure-ratings/nema-rated-enclosures.html (accessed 2025-07-16).
(23) *IEC 60204-1:2016*. https://webstore.iec.ch/en/publication/26037 (accessed 2025-07-16).
(24) *ISO 13485:2016*. ISO. https://www.iso.org/standard/59752.html (accessed 2025-07-16).
(25) *ISO 12100:2010*. ISO. https://www.iso.org/standard/51528.html (accessed 2025-07-16).
(26) *ISO 10218-1:2025*. ISO. https://www.iso.org/standard/73933.html (accessed 2025-07-16).
(27) *ANSI/RIA R15.06-2012 - Industrial Robots and Robot Systems - Safety Requirements (CONTAINS CORRIGENDUM)*. https://webstore.ansi.org/standards/ria/ansiriar15062012?srsltid=AfmBOooJDaU5kieJEv97E2z-VE6-ww77K3Zuhg9nIwN67J4LdoZuX1m0 (accessed 2025-07-16).

# Appendix I

Appendix I provides the electrical schematics that show a detail outline of the overall electrical design of the ACAT.

AUTOMATED CONTACT ANGLE TESTER

SPRING 2025

ENG 402 CAPSTONE DESIGN

ELECTRICAL SCHEMATICS

GROUP: CONNOR BURGESS & KYLE DOUIN

| DRAWN: | DATE: | SCHOOL: |
|---|---|---|
| CWB | 2/24/25 | UNIVERSITY OF SOUTHERN MAINE |

| DWG NUM: | SIZE: | TITLE: |
|---|---|---|
| 2025-1-1 | A3 | AUTOMATED CONTACT ANGLE TESTER |

| TEAM: | | TITLE PAGE |
|---|---|---|
| KYLE DOUIN CONNOR BURGESS | SCALE: NONE | DWG TYPE: ELECTRICAL | SHEET 1 OF 35 |

## Index

| PAGE | TITLE | REVISION |
|---|---|---|
| 1 | TITLE PAGE | 1 |
| 2 | INDEX (THIS PAGE) | 1 |
| 3 | SYMBOLS AND ABBREVIATIONS | 1 |
| 4 | B.O.M. - AC VOLTAGE ENCLOSURE | 1 |
| 5 | B.O.M. - MAIN ENCLOSURE PART 1 OF 2 | 1 |
| 6 | B.O.M. - MAIN ENCLOSURE PART 2 OF 2 | 1 |
| 7 | FUSE AND CABLE DESCRIPTION | 1 |
| 8 | HIGH VOLTAGE ENCLOSURE | 1 |
| 9 | SPARE | 1 |
| 10 | POWER DISTRIBUTION | 1 |
| 11 | SPARE | 1 |
| 12 | SAFETY CIRUIT | 1 |
| 13 | SPARE | 1 |
| 14 | X & Y AXIS DRIVES | 1 |
| 15 | Z - AXIS DRIVE | 1 |
| 16 | SPARE | 1 |
| 17 | PNEUMATIC AND LIGHT TOWER CONTROL | 1 |
| 18 | SPARE | 1 |
| 19 | WATER DISPENSER | 1 |
| 20 | SPARE | 1 |
| 21 | RPI HAT CONFIGURATION | 1 |
| 22 | SPARE | 1 |
| 23 | RPI POWER | 1 |
| 24 | SPARE | 1 |
| 25 | RPI INPUTS | 1 |

| PAGE | TITLE | REVISION |
|---|---|---|
| 26 | RPI OUPUTS | 1 |
| 27 | SPARE | 1 |
| 28 | OPERATOR CONTROL ENCLOSURE | 1 |
| 29 | SPARE | 1 |
| 30 | HIGH VOLTAGE ENCLOSURE LAYOUT | 1 |
| 31 | HIGH VOLTAGE ENCLOSURE CUTOUT | 1 |
| 32 | MAIN ENCLOSURE LAYOUT | 1 |
| 33 | MAIN ENCLOSURE CUTOUT | 1 |
| 34 | OPERATOR CONTROL ENCLOSURE LAYOUT | 1 |
| 35 | OPERATOR CONTROL ENCLOSURE CUTOUT | 1 |

DRAWN: CWB | DATE: 2/24/25 | SCHOOL: UNIVERSITY OF SOUTHERN MAINE
DWG NUM: 2025-1-2 | SIZE: A3 | TITLE: AUTOMATED CONTACT ANGLE TESTER — INDEX
TEAM: KYLE DOUIN, CONNOR BURGESS | SCALE: NONE | DWG TYPE: ELECTRICAL | SHEET 2 OF 35

---

## Symbols and Abbreviations

| DEVICE ABBREVIATIONS | DESCRIPTION |
|---|---|
| GND | GROUND |
| TB | TERMINAL BLOCK |
| CBL | CABLE |
| PWS | POWER SUPPLY |
| LT | INDICATOR LIGHT |
| FU | FUSE |
| RPI | RASPBERRY PI |
| RBH | RASPBERRY PI HAT |
| GBH | GPIO BREAKOUT HAT |
| LT | INDICATOR LIGHT |
| SR | SAFETY RELAY |
| CR | CONTROL RELAY |
| MCR | MASTER CONTROL RELAY |
| DR | DRIVE CONTROLLER |
| PB | PUSHBUTTON |
| ES | EMERGENCY STOP |
| PL | PLUG |
| DI | DOOR INTERLOCK |
| WP | WATER PUMP |

| DEVICE SYMBOLS | DESCRIPTION |
|---|---|
| ▢ | TERMINAL BLOCK |
| | LIGHT |
| | RELAY |
| | SOLENOID |
| | N.C. CONTACT |
| | N.O. CONTACT |
| | LIMIT SWITCH |
| | INDICATOR LIGHT |
| | FUSE |
| | N.C. MUSHROOM PUSH BUTTON |
| | N.C. PUSH BUTTON |
| | N.O. PUSH BUTTON |
| | TO & FROM LINE REFERENCE |
| | WIRE INTERSECTION |
| « | PLUG OR CONNECTOR |

WIRE NUMBERING

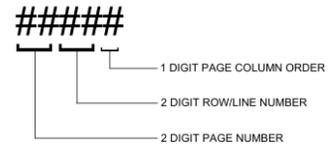

- 1 DIGIT PAGE COLUMN ORDER
- 2 DIGIT ROW/LINE NUMBER
- 2 DIGIT PAGE NUMBER

DEVICE AND CABLE NUMBERING

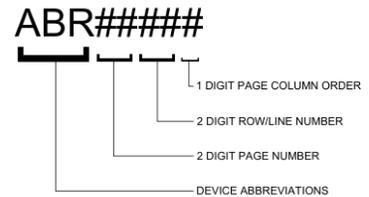

- 1 DIGIT PAGE COLUMN ORDER
- 2 DIGIT ROW/LINE NUMBER
- 2 DIGIT PAGE NUMBER
- DEVICE ABBREVIATIONS

DRAWN: CWB | DATE: 2/24/25 | SCHOOL: UNIVERSITY OF SOUTHERN MAINE
DWG NUM: 2025-1-3 | SIZE: A3 | TITLE: AUTOMATED CONTACT ANGLE TESTER — SYMBOLS AND ABBREVIATIONS
TEAM: KYLE DOUIN, CONNOR BURGESS | SCALE: NONE | DWG TYPE: ELECTRICAL | SHEET 3 OF 35

## AC VOLTAGE ENCLOSURE

| INDEX | DEV ID | MFG | MODEL | DESCRIPTION | QUANITY |
|---|---|---|---|---|---|
| 1 | | HOFFMAN | HOFFMAN | HIGH VOLTAGE ENCLOSURE | 1 |
| 2 | | HOFFMAN | HOFFMAN | HIGH VOLTAGE ENCLOSURE BACK PLATE | 1 |
| 3 | PL-08040 | HUBBLE | HBL5266C PLUG | MALE PLUG 15 AMP 125V | 1 |
| 4 | DSC-08040 | ABB | OS 30AJ12 | 3 POLE FUSED DISCONNECT; ONLY 1 POLE IS USED | 1 |
| 5 | FU-08040 | BUSSMAN | LPJ-4SP | 4AMP TIME DELAY CLASS J | 1 |
| 6 | TB-GND | PHOENIX CONTACT | HBL5266C PLUG | MALE PLUG 15 AMP 125V | 1 |
| 7 | PWS-08040 | MEANWELL | NDR-120-24 | 24 VDC POWER SUPPLY; 120W | 1 |
| 8 | PWS-08160 | TOBSUN | EA120-12V | 24VDC TO 12VDC STEP DOWN CONVERTER | 1 |
| 9 | | XHF | #2 - 3/8 STRAIN RELIEF | TWO STRAIN RELIEFS FOR CBL-08040 AND CBL-08041 | 1 |

DRAWN: CWB  DATE: 2/24/25  SCHOOL: UNIVERSITY OF SOUTHERN MAINE
DWG NUM: 2025-1-4  SIZE: A3  TITLE: AUTOMATED CONTACT ANGLE TESTER BILL OF MATERIALS AC VOLTAGE
TEAM: KYLE DOUIN, CONNOR BURGESS  SCALE: NONE  DWG TYPE: ELECTRICAL  SHEET 4 OF 35

## MAIN ENCLOSURE

| INDEX | DEV ID | MFG | MODEL | DESCRIPTION | QUANITY |
|---|---|---|---|---|---|
| 1 | | | HOFFMAN | MAIN ENCLOSURE | 1 |
| 2 | | | HOFFMAN | MAIN ENCLOSURE WIRE MANAGEMENT | 1 |
| 3 | CBL-08040 | | CABLE GLAN | 14 AWG 3 CONDUCTOR CABLE | 1 |
| 4 | SR-12050 | AB | MSR138DP | SAFETY RELAY | 1 |
| 5 | DI-12410 | OMRON | OS 30AJ12 | 3 POLE FUSED DISCONNECT; ONLY 1 POLE IS USED | 1 |
| 6 | ES-12530 | BUSSMAN | LPJ-4SP | 4AMP TIME DELAY CLASS J | 1 |
| 7 | ES-28190 | PHOENIX CONTACT | HBL5266C PLUG | MALE PLUG 15 AMP 125V | 1 |
| 8 | MCR-12590 | MEANWELL | NDR-120-24 | 24 VDC POWER SUPPLY; 120W | 1 |
| 9 | MCR-12620 | AB | EA120-12V | 24VDC TO 12VDC STEP DOWN CONVERTER | 1 |
| 10 | MCR-12590 | AB | #2 - 3/8 STRAIN RELIEF | TWO STRAIN RELIEFS FOR CBL-08040 AND CBL-08041 | 1 |
| 11 | MCR-12660 | AB | EA120-12V | 24VDC TO 12VDC STEP DOWN CONVERTER | 1 |
| 12 | MCR-12690 | AB | #2 - 3/8 STRAIN RELIEF | TWO STRAIN RELIEFS FOR CBL-08040 AND CBL-08041 | 1 |
| 13 | DR-14010 | | HOFFMAN | MAIN ENCLOSURE | 1 |
| 14 | DR-14410 | | HOFFMAN | MAIN ENCLOSURE WIRE MANAGEMENT | 1 |
| 15 | DR-15040 | | CABLE GLAN | 14 AWG 3 CONDUCTOR CABLE | 1 |
| 16 | LT-17250 | AB | MSR138DP | SAFETY RELAY | 1 |
| 17 | DR-19040 | OMRON | OS 30AJ12 | 3 POLE FUSED DISCONNECT; ONLY 1 POLE IS USED | 1 |
| 18 | CR-19420 | BUSSMAN | LPJ-4SP | 4AMP TIME DELAY CLASS J | 1 |
| 19 | WP-1948 | PHOENIX CONTACT | HBL5266C PLUG | MALE PLUG 15 AMP 125V | 1 |
| 20 | PL-21100 | PATELITE | NDR-120-24 | 24 VDC POWER SUPPLY; 120W | 1 |
| 21 | PL-21120 | AB | EA120-12V | 24VDC TO 12VDC STEP DOWN CONVERTER | 1 |
| 22 | PL-21230 | AB | #2 - 3/8 STRAIN RELIEF | TWO STRAIN RELIEFS FOR CBL-08040 AND CBL-08041 | 1 |
| 23 | RPI-21070 | AB | EA120-12V | 24VDC TO 12VDC STEP DOWN CONVERTER | 1 |
| 24 | RFH-21440 | AB | #2 - 3/8 STRAIN RELIEF | TWO STRAIN RELIEFS FOR CBL-08040 AND CBL-08041 | 1 |
| 25 | PS-25100 | AB | EA120-12V | 24VDC TO 12VDC STEP DOWN CONVERTER | 1 |
| 26 | PS-25140 | AB | #2 - 3/8 STRAIN RELIEF | TWO STRAIN RELIEFS FOR CBL-08040 AND CBL-08041 | 1 |
| 27 | LS-25480 | AB | EA120-12V | 24VDC TO 12VDC STEP DOWN CONVERTER | 1 |
| 28 | LS-25520 | AB | #2 - 3/8 STRAIN RELIEF | TWO STRAIN RELIEFS FOR CBL-08040 AND CBL-08041 | 1 |

DRAWN: CWB  DATE: 2/24/25  SCHOOL: UNIVERSITY OF SOUTHERN MAINE
DWG NUM: 2025-1-5  SIZE: A3  TITLE: AUTOMATED CONTACT ANGLE TESTER BILL OF MATERIALS MAIN ENCLOSURE PART 1 OF 2
TEAM: KYLE DOUIN, CONNOR BURGESS  SCALE: NONE  DWG TYPE: ELECTRICAL  SHEET 5 OF 35

## MAIN ENCLOSURE

| INDEX | DEV ID | MFG | MODEL | DESCRIPTION | QUANITY |
|---|---|---|---|---|---|
| 29 | CR-26220 | AEDIK | AE-19523 | 1 CHANNEL RELAY MODULE | 1 |
| 30 | CR-26510 | AEDIK | AE-19523 | 1 CHANNEL RELAY MODULE | 1 |
| 31 | CR-26530 | AEDIK | AE-19523 | 1 CHANNEL RELAY MODULE | 1 |
| 32 | CR-26550 | AEDIK | AE-19523 | 1 CHANNEL RELAY MODULE | 1 |
| 33 | CR-26551 | AEDIK | AE-19523 | 1 CHANNEL RELAY MODULE | 1 |
| 34 | CR-26590 | AEDIK | AE-19523 | 1 CHANNEL RELAY MODULE | 1 |
| 35 | CR-26591 | AEDIK | AE-19523 | 1 CHANNEL RELAY MODULE | 1 |
| 36 |  |  |  |  | 1 |

## OPERATOR CONTROL ENCLOSURE

| INDEX | DEV ID | MFG | MODEL | DESCRIPTION | QUANITY |
|---|---|---|---|---|---|
| 1 |  | HOFFMAN |  | MAIN ENCLOSURE | 1 |
| 2 |  | HOFFMAN |  | MAIN ENCLOSURE WIRE MANAGEMENT | 1 |
| 3 |  |  | CABLE GLAN | 14 AWG 3 CONDUCTOR CABLE | 1 |
| 4 | PB-28050 | AB | MSR138DP | SAFETY RELAY | 1 |
| 5 | PB-28130 | AB | OS 30AJ12 | 3 POLE FUSED DISCONNECT; ONLY 1 POLE IS USED | 1 |
| 6 | ES-28150 | BUSSMAN | LPJ-4SP | 4AMP TIME DELAY CLASS J | 1 |
| 7 |  |  |  |  | 1 |

---

DRAWN: CWB  DATE: 2/24/25  SCHOOL: UNIVERSITY OF SOUTHERN MAINE
DWG NUM: 2025-1-6  SIZE: A3  TITLE: AUTOMATED CONTACT ANGLE TESTER BILL OF MATERIALS MAIN ENCLOSURE PART 2 OF 2
TEAM: KYLE DOUIN, CONNOR BURGESS
SCALE: NONE  DWG TYPE: ELECTRICAL  SHEET 6 OF 35

---

## AC VOLTAGE ENCLOSURE : FUSES

| FUSE | RATING | MFG | MODEL | DESCRIPTION | QUANITY |
|---|---|---|---|---|---|
| FU-08040 | 4.0 AMPS | BUSSMAN | LPJ-4SP | CLASS J, POWER CIRCUIT | 1 |

## AC VOLTAGE ENCLOSURE : CABLES

| CABLE | AWG | COUNT | MFG | MODEL | DESCRIPTION | LENGTH |
|---|---|---|---|---|---|---|
| CBL-08040 | 14 AWG | 3 | SOUTHWIRE | 55809943 | POWER CIRCUIT | 4' |
| CBL-08041 | 18 AWG | 7 | MFG | MODEL | POWER CIRCUIT | 6' |

## MAIN ENCLOSURE : FUSES

| FUSE | RATING | MFG | MODEL | DESCRIPTION | QUANITY |
|---|---|---|---|---|---|
| FU-10100 | 0.5 AMPS | HUAREW | HRF-FG24V-240P | GLASS, 5MM X 20MM, FAN MOTOR | 1 |
| FU-10190 | 0.5 AMPS | BUSSMAN | HRF-FG24V-240P | GLASS, 5MM X 20MM, SAFETY CIRCUIT | 1 |
| FU-10220 | 1.0 AMPS | BUSSMAN | HRF-FG24V-240P | GLASS, 5MM X 20MM, X-AXIS DRIVE | 1 |
| FU-10250 | 1.0 AMPS | BUSSMAN | HRF-FG24V-240P | GLASS, 5MM X 20MM, Y-AXIS DRIVE | 1 |
| FU-10280 | 1.0 AMPS | BUSSMAN | HRF-FG24V-240P | GLASS, 5MM X 20MM, Z-AXIS DRIVE | 1 |
| FU-10310 | 0.5 AMPS | BUSSMAN | HRF-FG24V-240P | GLASS, 5MM X 20MM, W.D. DRIVE | 1 |
| FU-10340 | 0.5 AMPS | BUSSMAN | HRF-FG24V-240P | GLASS, 5MM X 20MM, RBH POWER | 1 |
| FU-10450 | 1.0 AMPS | BUSSMAN | HRF-FG24V-240P | GLASS, 5MM X 20MM, ALL PURPOSE 24V | 1 |
| FU-19480 | 1.6 AMPS | BUSSMAN | HRF-FG24V-240P | GLASS, 5MM X 20MM, WATER PUMP | 1 |

## MAIN ENCLOSURE : CABLES

| CABLE | AWG | COUNT | MFG | MODEL | DESCRIPTION | LENGTH |
|---|---|---|---|---|---|---|
| CBL-12290, CBL-12430, CBL-17251, CBL-19480 | 14 AWG | 4 | MAX BRITE | 184-100FS | USED FOR ESTOPS & LIGHT INDICATOR | 6' |
| CBL-14121, CBL-14491, CBL-15211, CBL-19121 | 24 AWG | 8 | BLACK BOX | EVNSL601A-1000 | USED FOR STEPPER MOTORS | 6' |
| CBL-21100, CBL-21120, CBL-21130 | USB A TO USB A |  | CONABLE | CAL3SMM-3FT-3PK | USED RPI COMMUNICATION | 3' |
| CBL-21230 | HDMI TO MICRO HDMI |  | AMAZON | HDMI-AD-6FT | USED RPI COMMUNICATION | 6' |

## OPERATOR CONTROL ENCLOSURE : CABLES

| CABLE | AWG | COUNT | MFG | MODEL | DESCRIPTION | LENGTH |
|---|---|---|---|---|---|---|
| CBL-28050 | 18 AWG | 7 | AMAZON | HDMI-AD-6FT | USED OPERATOR CONTROL ENCLOSURE | 8' |

---

DRAWN: CWB  DATE: 2/24/25  SCHOOL: UNIVERSITY OF SOUTHERN MAINE
DWG NUM: 2025-1-7  SIZE: A3  TITLE: AUTOMATED CONTACT ANGLE TESTER FUSE AND CABLE DESCRIPTION
TEAM: KYLE DOUIN, CONNOR BURGESS
SCALE: NONE  DWG TYPE: ELECTRICAL  SHEET 7 OF 35

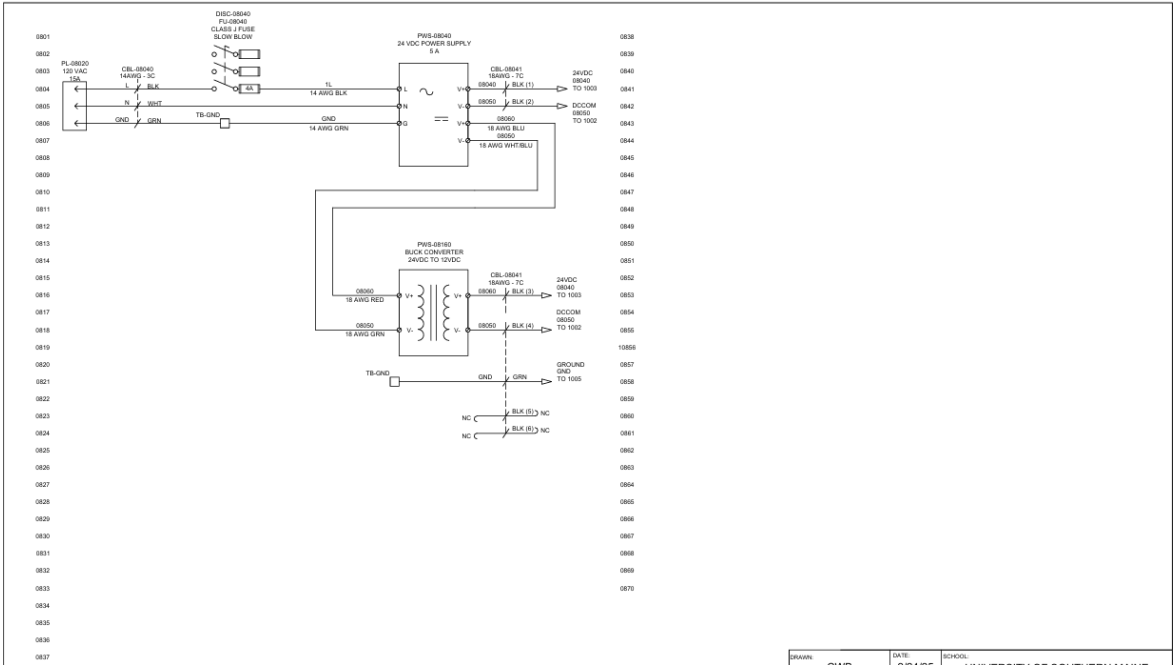

# INTENTIONALLY BLANK

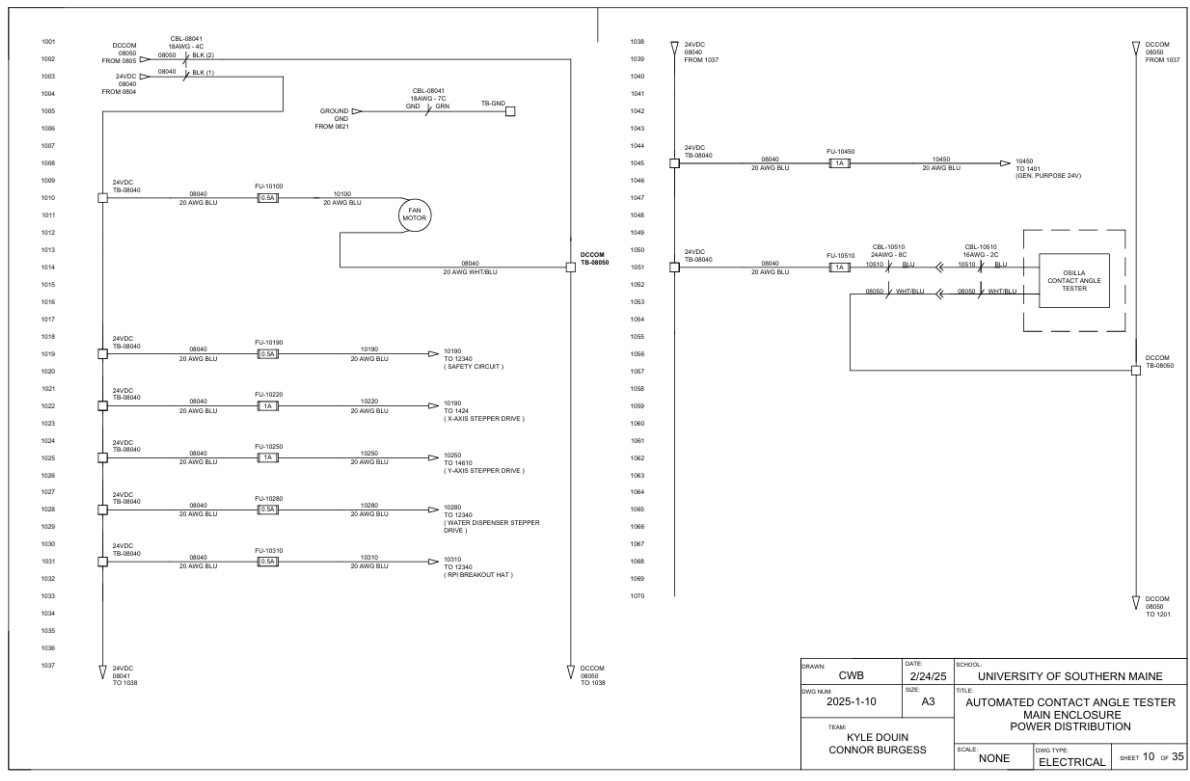

# INTENTIONALLY BLANK

Sheet 11 of 35 — SPARE — Automated Contact Angle Tester — University of Southern Maine — Drawn: CWB, 2/24/25 — DWG NUM: 2025-1-11 — Size A3 — Team: Kyle Douin, Connor Burgess — Scale: None — DWG Type: Electrical

---

Sheet 12 of 35 — Automated Contact Angle Tester — Main Enclosure Safety Circuit — University of Southern Maine — Drawn: CWB, 2/24/25 — DWG NUM: 2025-1-12 — Size A3 — Team: Kyle Douin, Connor Burgess — Scale: None — DWG Type: Electrical

# INTENTIONALLY BLANK

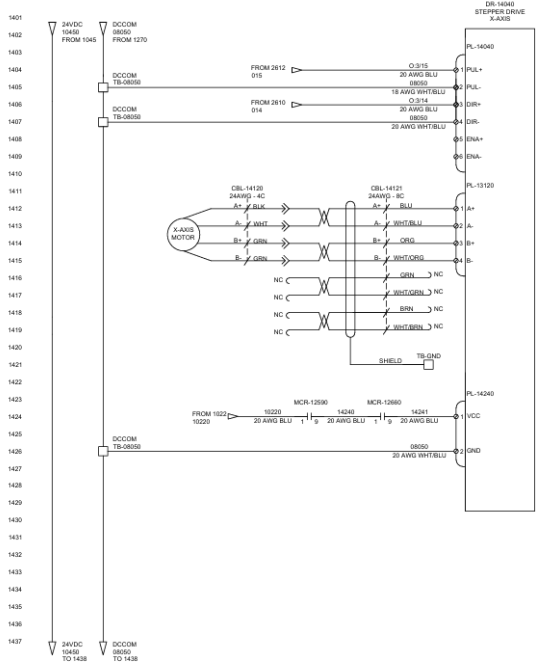
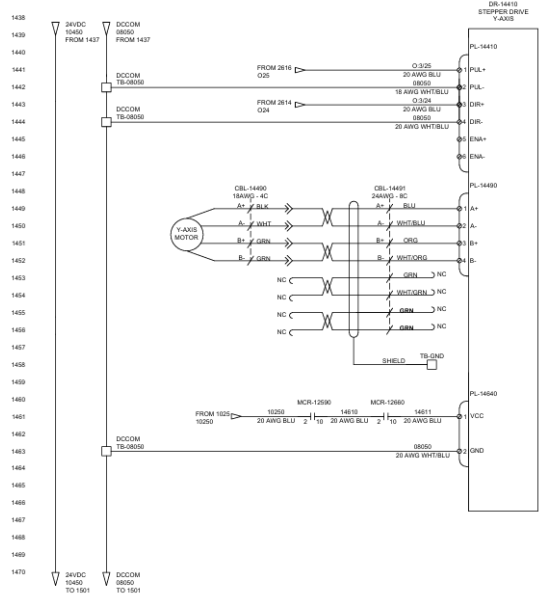

INTENTIONALLY BLANK

INTENTIONALLY BLANK

Sheet 17 of 35 — Automated Contact Angle Tester, Main Enclosure, Pneumatic and Light Tower Control (DWG 2025-1-17, CWB, 2/24/25, Team: Kyle Douin / Connor Burgess).

# INTENTIONALLY BLANK

Sheet 18 of 35 — Automated Contact Angle Tester, Spare (DWG 2025-1-18, CWB, 2/24/25, Team: Kyle Douin / Connor Burgess).

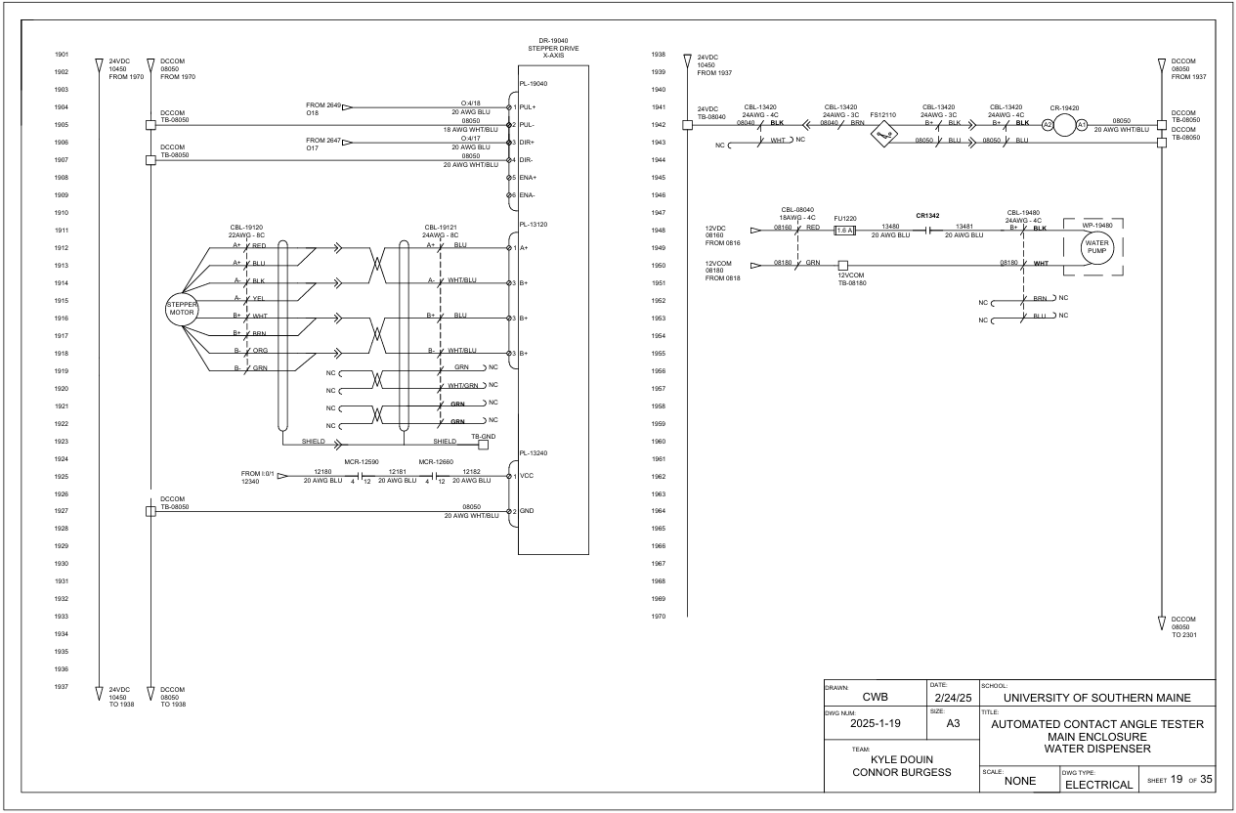

INTENTIONALLY BLANK

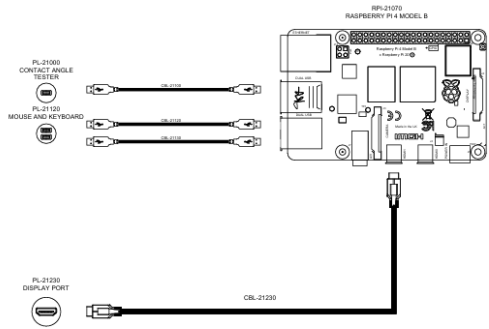
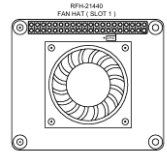
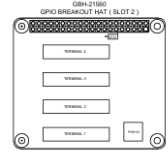

| | | | |
|---|---|---|---|
| DRAWN: CWB | DATE: 2/24/25 | SCHOOL: UNIVERSITY OF SOUTHERN MAINE | |
| DWG NUM: 2025-1-21 | SIZE: A3 | TITLE: AUTOMATED CONTACT ANGLE TESTER MAIN ENCLOSURE RPI HAT CONFIGURATION | |
| TEAM: KYLE DOUIN CONNOR BURGESS | | SCALE: NONE | DWG TYPE: ELECTRICAL | SHEET 21 OF 35 |

# INTENTIONALLY BLANK

| | | | |
|---|---|---|---|
| DRAWN: CWB | DATE: 2/24/25 | SCHOOL: UNIVERSITY OF SOUTHERN MAINE | |
| DWG NUM: 2025-1-22 | SIZE: A3 | TITLE: AUTOMATED CONTACT ANGLE TESTER SPARE | |
| TEAM: KYLE DOUIN CONNOR BURGESS | | SCALE: NONE | DWG TYPE: ELECTRICAL | SHEET 22 OF 35 |

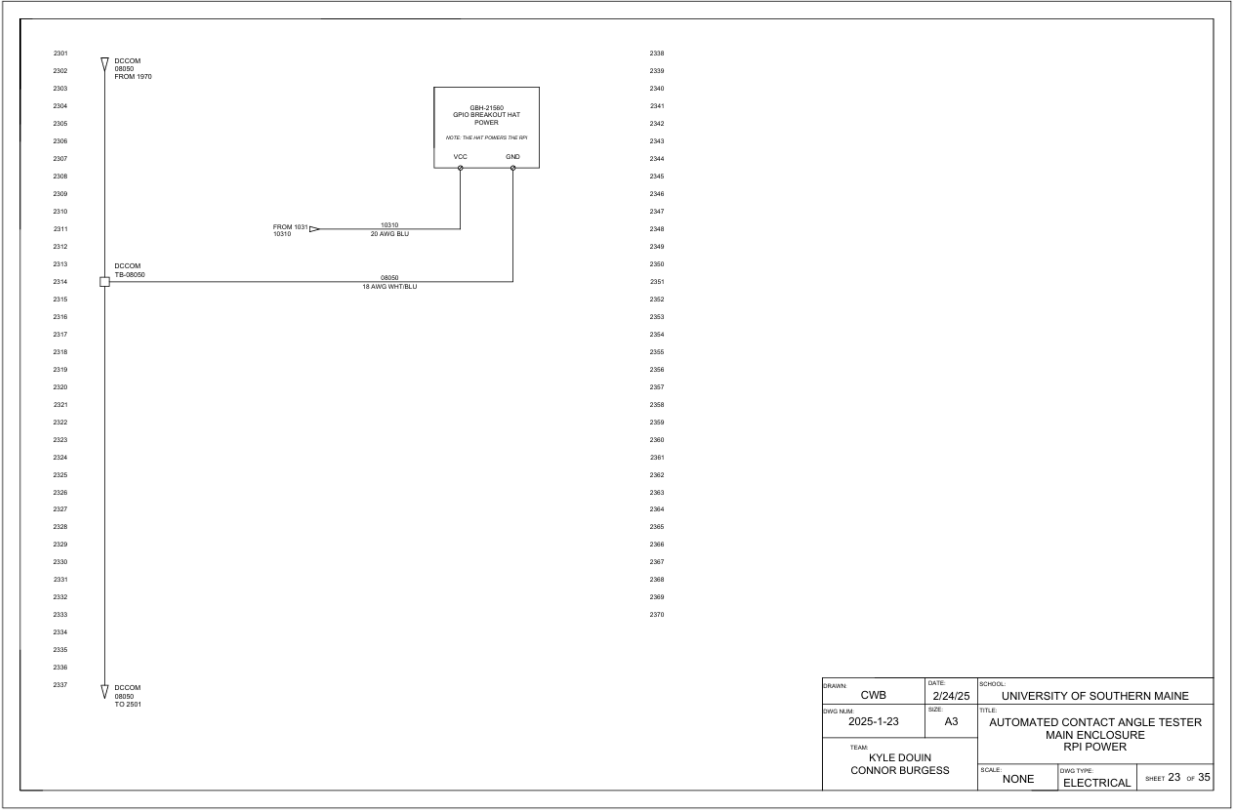

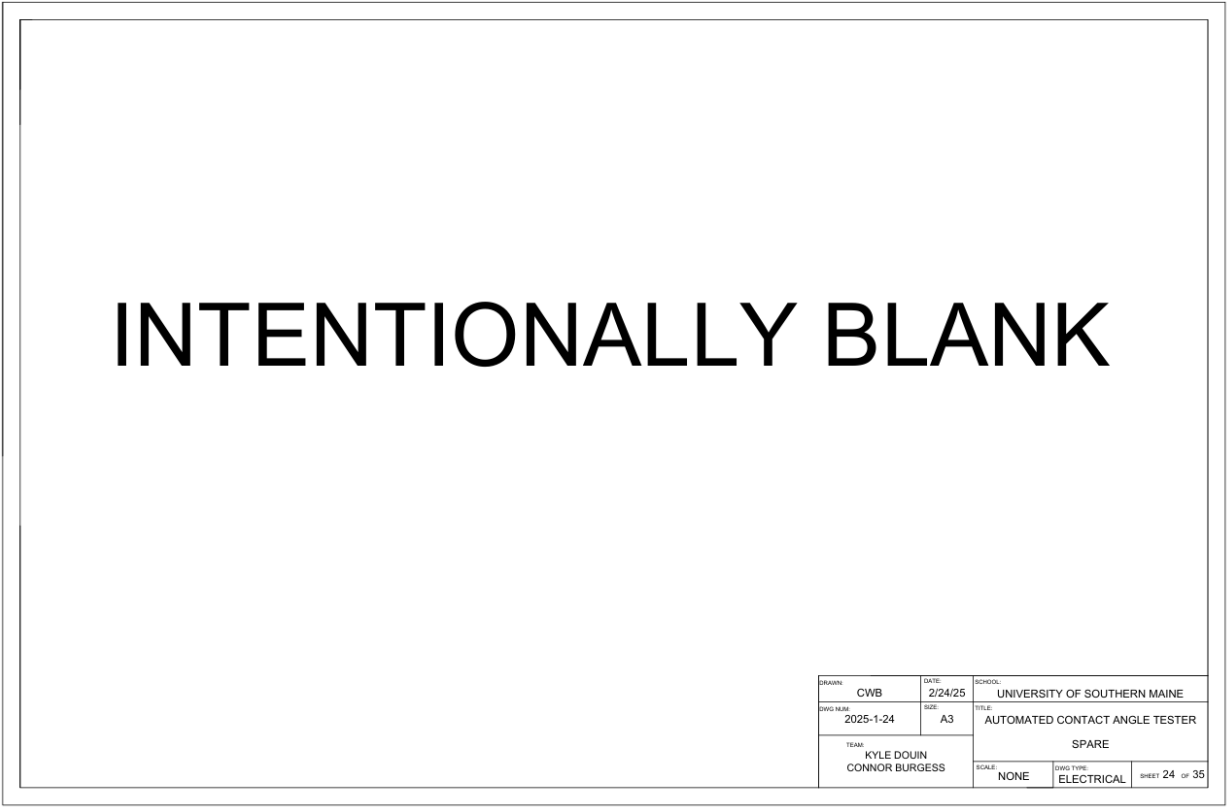

INTENTIONALLY BLANK

# INTENTIONALLY BLANK

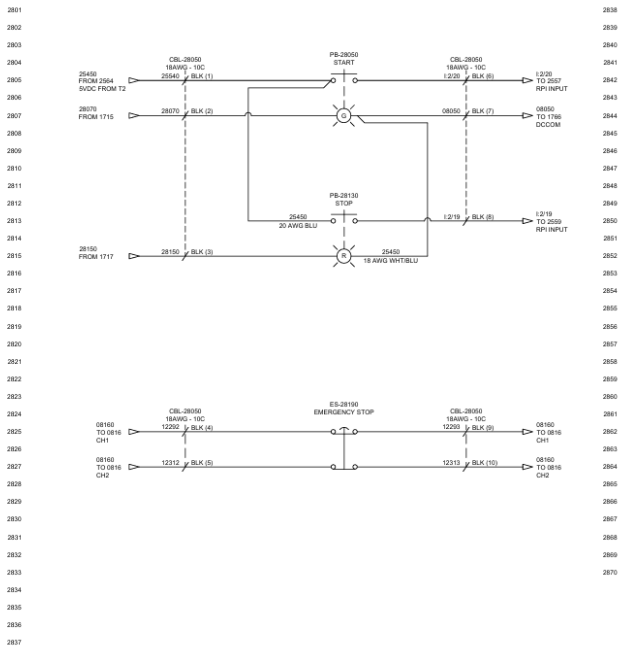

# INTENTIONALLY BLANK

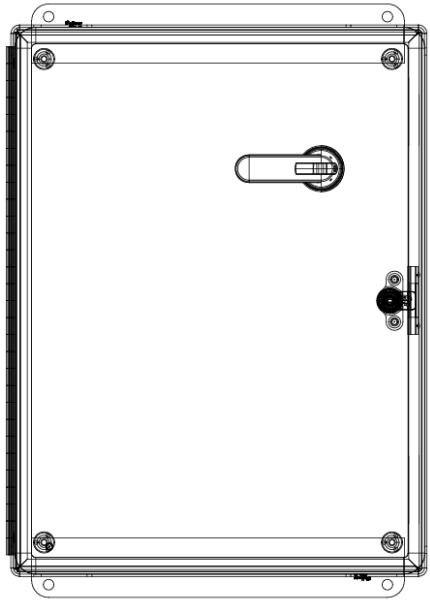
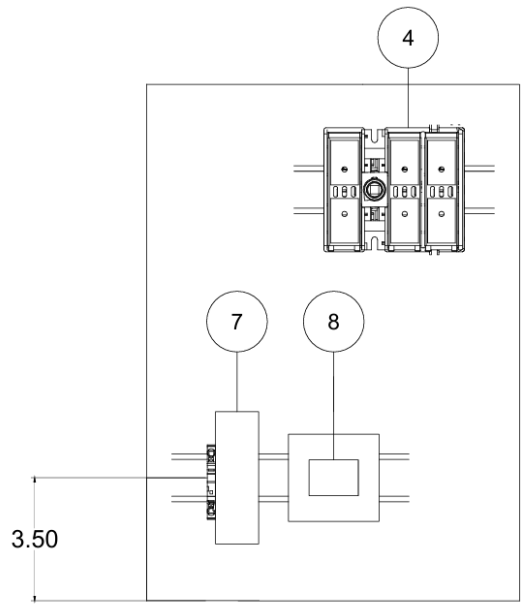

3.50

## Sheet 31 of 35 — AC Enclosure Panel Cutout

- 3.0000
- Ø1.0000
- 4.0000

| DRAWN: | DATE: | SCHOOL: |
|---|---|---|
| CWB | 2/24/25 | UNIVERSITY OF SOUTHERN MAINE |
| DWG NUM: | SIZE: | TITLE: |
| 2025-1-31 | A3 | AUTOMATED CONTACT ANGLE TESTER<br>AC ENCLOSURE<br>PANEL CUTOUT |
| TEAM: | | |
| KYLE DOUIN<br>CONNOR BURGESS | SCALE:<br>1" = 2" | DWG TYPE:<br>MECHANICAL · SHEET 31 OF 35 |

## Sheet 32 of 35 — Main Enclosure Panel Layout

Callouts on panel: 13, 14, 15, 20, 21, 22, 23, 24, 26, 2, 5, 6, TB-1, TB-2, 7, 8, 11, 29-37

**TB-2**
ROTATED 90 DEGREES CLOCKWISE
NUMBERS ARE IDENTICAL TO THEIR
CORRESPONDING TERMINAL BLOCK

| TB-2 |
|---|
| 25080 |
| 25450 |
| 10190 |
| 10450 |
| 10450 |
| 10450 |
| 08050 |
| 08050 |
| 08050 |
| 08050 |
| 08050 |
| 08050 |
| 08050 |
| 08050 |
| 08050 |
| 08040 |
| 08040 |
| 08040 |
| GND |
| GND |
| GND |

**TB-1**
ROTATED 90 DEGREES CLOCKWISE
NUMBERS ARE IDENTICAL TO THEIR
CORRESPONDING SIDE

| TOP | BOTTOM |
|---|---|
| 12291 | 12292 |
| 12311 | 12312 |
| SPARE | SPARE |
| SPARE | SPARE |
| SPARE | SPARE |

| DRAWN: | DATE: | SCHOOL: |
|---|---|---|
| CWB | 2/24/25 | UNIVERSITY OF SOUTHERN MAINE |
| DWG NUM: | SIZE: | TITLE: |
| 2025-1-32 | A3 | AUTOMATED CONTACT ANGLE TESTER<br>MAIN ENCLOSURE<br>PANEL LAYOUT |
| TEAM: | | |
| KYLE DOUIN<br>CONNOR BURGESS | SCALE:<br>1" = 4" | DWG TYPE:<br>MECHANICAL · SHEET 32 OF 35 |

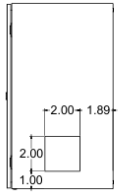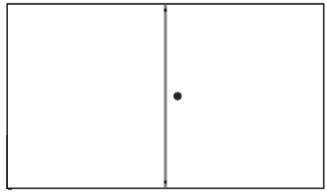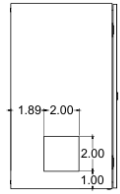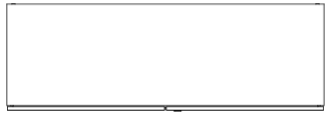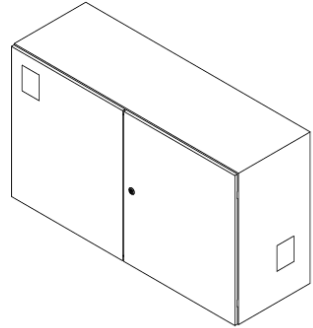

| DRAWN: CWB | DATE: 2/24/25 | SCHOOL: UNIVERSITY OF SOUTHERN MAINE |
| --- | --- | --- |
| DWG NUM: 2025-1-33 | SIZE: A3 | TITLE: AUTOMATED CONTACT ANGLE TESTER MAIN ENCLOSURE PANEL CUTOUT |
| TEAM: KYLE DOUIN CONNOR BURGESS | SCALE: 1" = 4" | DWG TYPE: MECHANICAL — SHEET 33 OF 35 |

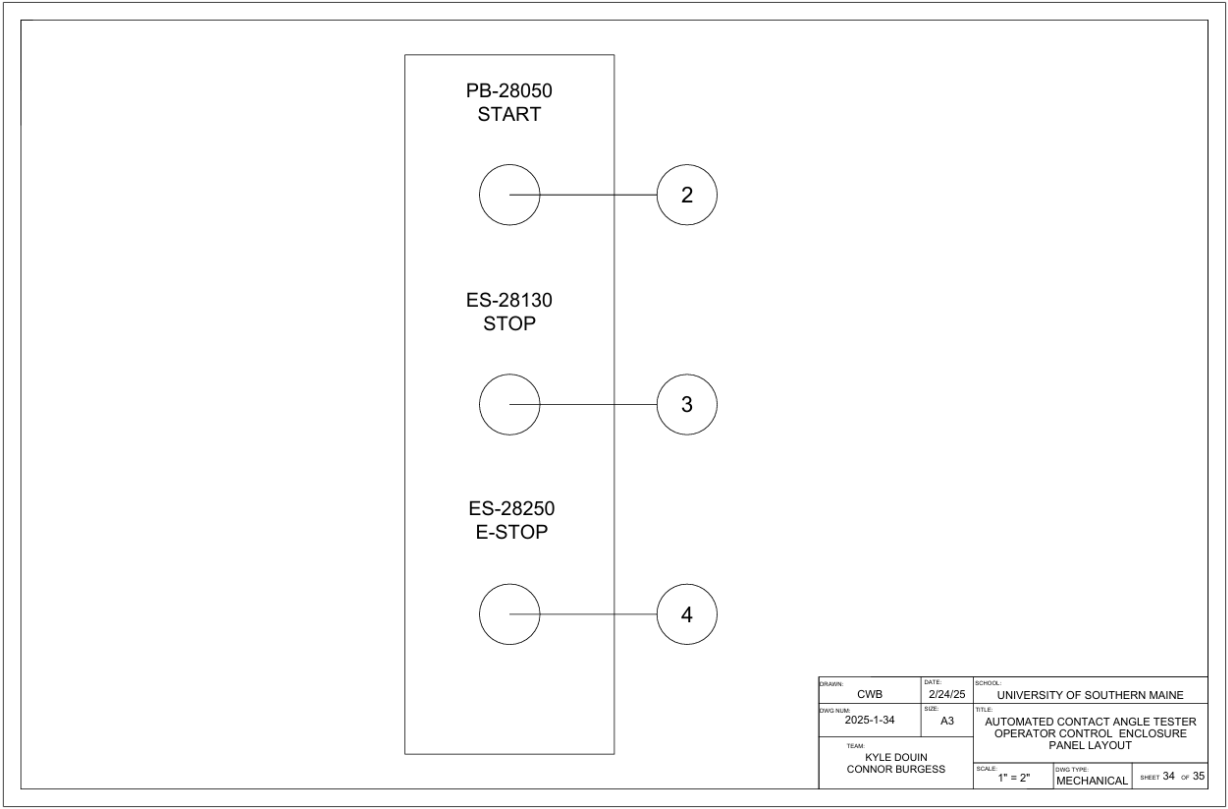
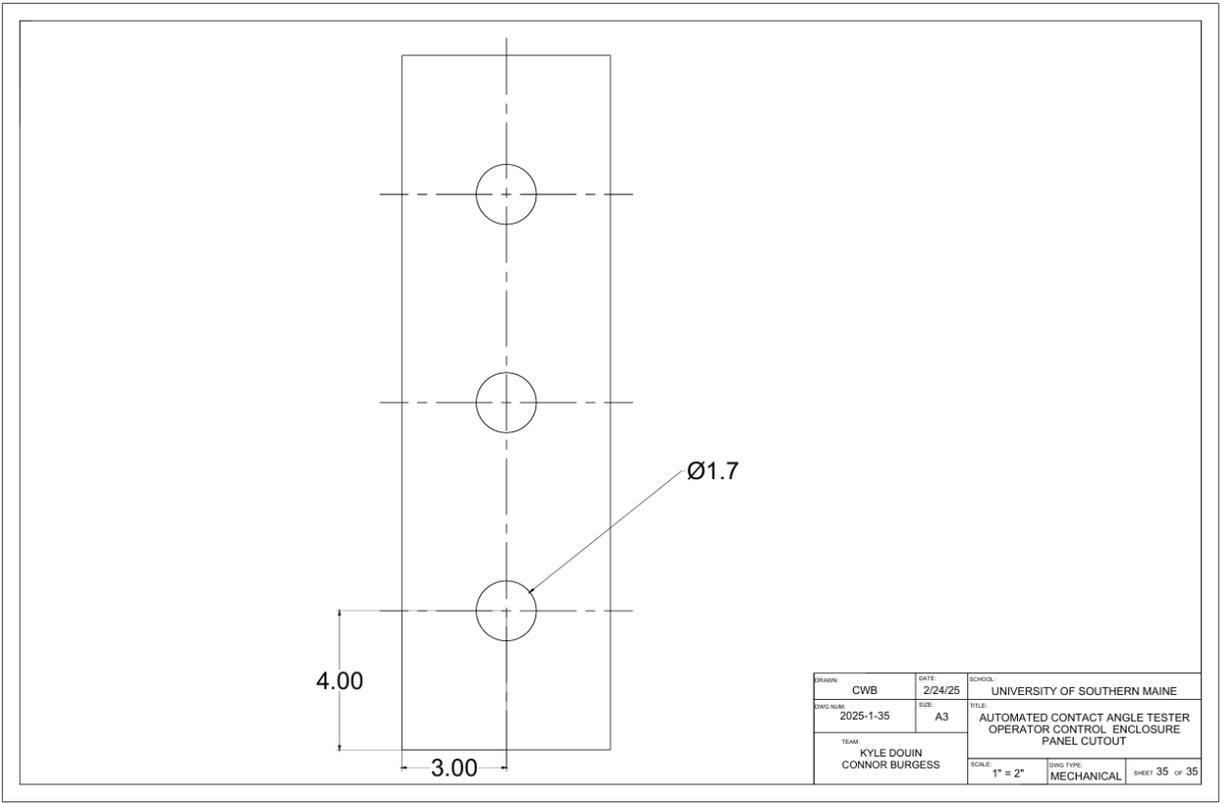